\documentclass[nohyperref]{article}

\usepackage{microtype}
\usepackage{graphicx}
\usepackage{subfigure}
\usepackage{booktabs} 
\usepackage{units}
\usepackage{placeins}

\usepackage{hyperref}

\usepackage[accepted]{icml2022}

\usepackage{amsmath}
\usepackage{amssymb}
\usepackage{mathtools}
\usepackage{amsthm}
\usepackage{arydshln}

\usepackage[capitalize,noabbrev]{cleveref}

\theoremstyle{plain}

\theoremstyle{definition}

\theoremstyle{remark}

\newcommand{\papertitle}{On the Difficulty of Defending Self-Supervised Learning against Model Extraction}
\icmltitlerunning{\papertitle}

\usepackage{amsmath,amsfonts,bm}

\def\eqref#1{equation~\ref{#1}}

\def\1{\bm{1}}

\DeclareMathAlphabet{\mathsfit}{\encodingdefault}{\sfdefault}{m}{sl}
\SetMathAlphabet{\mathsfit}{bold}{\encodingdefault}{\sfdefault}{bx}{n}

\usepackage{xparse}
\usepackage{xspace}
\usepackage{algorithm}
\usepackage{fixltx2e}
\usepackage{amsmath}
\usepackage{tikz}
\usepackage{float}
\usepackage{multirow}
\usepackage{multicol}
\usepackage{colortbl}

\usepackage{microtype}
\usepackage{graphicx}
\usepackage{subfigure}
\usepackage{booktabs} 
\usepackage[shortlabels]{enumitem}

\usepackage{hyperref,amssymb,enumitem}

\usepackage{wrapfig}

\setlist[itemize]{leftmargin=*}
\setlist[enumerate]{leftmargin=*}

\renewcommand{\algorithmicrequire}{\textbf{Input: }}

\renewcommand{\algorithmicensure}{\textbf{Output: }}

\usepackage{amsmath}

\DeclareRobustCommand\encircle[1]{\tikz[baseline=(char.base)]{\node[shape=circle,fill,inner sep=1pt] (char) {\textcolor{white}{#1}}}}

\newcommand{\augpredict}{\textit{augmentation predictor}\xspace}

\graphicspath{{images/}}

\newif\ifdraft
\draftfalse

\usepackage{fancyvrb}

\RecustomVerbatimCommand{\VerbatimInput}{VerbatimInput}%
{fontsize=\footnotesize,
 frame=lines,  
 framesep=2em, 
 commandchars=\|\(\), 
 commentchar=*        
}

\makeatletter
\def\adl@drawiv#1#2#3{%
        \hskip.5\tabcolsep
        \xleaders#3{#2.5\@tempdimb #1{1}#2.5\@tempdimb}%
                #2\z@ plus1fil minus1fil\relax
        \hskip.5\tabcolsep}
\newcommand{\cdashlinelr}[1]{%
  \noalign{\vskip\aboverulesep
           \global\let\@dashdrawstore\adl@draw
           \global\let\adl@draw\adl@drawiv}
  \cdashline{#1}
  \noalign{\global\let\adl@draw\@dashdrawstore
           \vskip\belowrulesep}}
\makeatother

\ifdraft
\newcommand{\mynote}[1]{\textcolor{red}{[note: #1]}}

\newcommand{\mytodo}[1]{\textcolor{red}{[todo: #1]}}
\newcommand{\mycomment}[1]{\textcolor{red}{[comment: #1]}}
\newcommand{\chris}[1]{\textcolor{red}{Chris: #1}}
\newcommand{\ahmad}[1]{\textcolor{blue}{Ahmad: #1}}
\newcommand{\vinith}[1]{\textcolor{blue}{Vinith: #1}}
\newcommand{\yunxiang}[1]{\textcolor{cyan}{Yunxiang: #1}}
\newcommand{\xiao}[1]{\textcolor{blue}{xiao: #1}}
\definecolor{chocolate(traditional)}{rgb}{0.48, 0.25, 0.0}

\definecolor{darkpastelgreen}{rgb}{0.01, 0.75, 0.24}
\newcommand{\natalie}[1]{\textcolor{darkpastelgreen}{natalie: #1}}
\definecolor{pistachio}{rgb}{0.58, 0.77, 0.45}

\newcommand{\jonas}[1]{\textcolor{magenta}{[Jonas: #1]}}

\newcommand{\adelin}[1]{\textcolor{red}{[Adelin: #1]}}
\newcommand{\mohammad}[1]{\textcolor{red}{[Mohammad: #1]}}

\definecolor{amber(sae/ece)}{rgb}{1.0, 0.49, 0.0}
\newcommand\adam[1]{{\textcolor{red}{[Adam: #1]}}}
\newcommand{\sierra}[1]{\textcolor{blue}{[Sierra: #1]}}
\newcommand{\armin}[1]{\textcolor{cyan}{[Armin: #1]}}
\newcommand{\nikita}[1]{\textcolor{cyan}{[Nikita: #1]}}

\else

\newcommand{\chris}[1]{}
\newcommand{\vinith}[1]{}
\newcommand{\adam}[1]{}
\newcommand{\yunxiang}[1]{}
\newcommand{\natalie}[1]{}
\newcommand{\jonas}[1]{}
\newcommand{\adelin}[1]{}
\newcommand{\mynote}[1]{}
\newcommand{\xiao}[1]{}
\newcommand{\mytodo}[1]{}
\newcommand{\mycomment}[1]{}
\newcommand{\ahmad}[1]{}
\newcommand{\mohammad}[1]{}
\newcommand{\sierra}[1]{}
\newcommand{\armin}[1]{}
\newcommand{\nikita}[1]{}
\fi

\usepackage{tikz}
\usetikzlibrary{mindmap}
\usepackage[linguistics]{forest}

\DeclareRobustCommand\encircle[1]{\tikz[baseline=(char.base)]{\node[shape=circle,fill,inner sep=1pt] (char) {\textcolor{white}{#1}}}}

\begin{document}

\twocolumn[
\icmltitle{\papertitle}




\begin{icmlauthorlist}
\icmlauthor{Adam Dziedzic}{uoft,vector}
\icmlauthor{Nikita Dhawan}{uoft,vector}
\icmlauthor{Muhammad Ahmad Kaleem}{uoft,vector}
\icmlauthor{Jonas Guan}{uoft,vector}
\icmlauthor{Nicolas Papernot}{uoft,vector}
\end{icmlauthorlist}

\icmlaffiliation{uoft}{University of Toronto}
\icmlaffiliation{vector}{Vector Institute}

\icmlcorrespondingauthor{Adam Dziedzic}{adam.dziedzic@utoronto.ca}

\icmlkeywords{Machine Learning, Model Stealing and Defenses, Representation Learning}

\vskip 0.3in
]



\printAffiliationsAndNotice{}  

\begin{abstract}
Self-Supervised Learning (SSL) is an increasingly popular ML paradigm that trains models to transform complex inputs into representations without relying on explicit labels. These representations encode similarity structures that enable efficient learning of multiple downstream tasks. Recently, ML-as-a-Service providers have commenced offering trained SSL models over inference APIs, which transform user inputs into useful representations for a fee. However, the high cost involved to train these models and their exposure over APIs both make black-box extraction a realistic security threat. We thus explore model stealing attacks against SSL. Unlike traditional model extraction on classifiers that output labels, the victim models here output representations; these representations are of significantly higher dimensionality compared to the low-dimensional prediction scores output by classifiers. We construct several novel attacks and find that approaches that train directly on a victim's stolen representations are query efficient and enable high accuracy for downstream models. We then show that existing defenses against model extraction are inadequate and not easily retrofitted to the specificities of SSL.

\end{abstract}

\section{Introduction}

Self-Supervised Learning (SSL) trains encoder models to transform unlabeled inputs into useful representations that are amenable to sample-efficient learning of multiple downstream tasks. 
The ability of SSL models to learn from unlabeled data has made them increasingly popular in important domains like computer vision, natural language processing, and speech recognition, where data are often abundant but labeling them is expensive~\citep{simclr, byol, moco, clip}.
Recently, ML-as-a-Service providers like OpenAI~\citep{neelakantan2022text, OpenAI} and Cohere~\citep{cohere} have begun offering trained encoders over inference APIs. 
Users pay a fee to transform their input data into intermediate representations, which are then used for downstream models.
High-performance SSL models in these domains are often costly to train; training a large BERT model can cost north of 1 million USD~\citep{sharir2020cost}.
The value of these models and their exposure over publicly-accessible APIs make black-box model extraction attacks a realistic security threat.

In a model extraction attack~\cite{pred_apis}, the attacker aims to steal a copy of the victim's model by submitting carefully selected queries and observing the outputs.
The attacker uses the query-output pairs to rapidly train a local model, often at a fraction of the cost of the victim model~\citep{fidelity}.
The stolen model can be used for financial gains or as a
reconnaissance stage to mount further attacks
like extracting private training data or constructing adversarial examples~\citep{papernot2017practical}.

Past work on model extraction focused on the Supervised Learning (SL) setting, where the victim model typically returns a label or other low-dimensional outputs like confidence scores~\cite{pred_apis} or logits~\citep{DataFreeExtract}.
In contrast, SSL encoders return high-dimensional representations; the de facto output for a ResNet-50 SimCLR model, a popular architecture in vision, is a 2048-dimensional vector.
We hypothesize this significantly higher information leakage from encoders makes them more vulnerable to extraction attacks than SL models.

In this paper, we introduce and compare several novel encoder extraction attacks to empirically demonstrate their threat to SSL models, and discuss the difficulties in protecting such models with existing defenses.

The framework of our attacks is inspired by Siamese networks: we query the victim model with inputs from a reference dataset similar to the victim's training set (e.g. CIFAR10 against an ImageNet victim), and then use the query-output pairs to train a local encoder to output the same representations as the victim.
We use two methods to train the local model during extraction.
The first method directly minimizes the loss between the output layer of the local model and the output of the victim.
The second method utilizes a separate projection head, which recreates the contrastive learning setup utilized by SSL frameworks like SimCLR.
For each method, we compare different loss metrics, including MSE, InfoNCE, Soft Nearest Neighbors (SoftNN), or Wasserstein distance.

We evaluate our attacks against ResNet-50 and ResNet-34 SimCLR victim models. 
We train the models and generate queries for the extraction attacks using ImageNet and CIFAR10, and evaluate the models' downstream performance on CIFAR100, STL10, SVHN, and Fashion-MNIST.
We also simulate attackers with different levels of access to data that is in or out of distribution \textit{w.r.t.} the victim's training data.

Our experimental results in Section~\ref{experiments} show that our attacks can steal a copy of the victim model that achieves considerable downstream performance in fewer than $\nicefrac{1}{5}$ of the queries used to train the victim. 
Against a victim model trained on 1.2M unlabeled samples from ImageNet, with a 91.9\% accuracy on the downstream Fashion-MNIST classification task, our direct extraction attack with the InfoNCE loss stole a copy of the encoder that achieves 90.5\% accuracy in 200K queries.
Similarly, against a victim trained on 50K unlabeled samples from CIFAR10, with a 79.0\% accuracy on the downstream CIFAR10 classification task, our direct extraction attack with the SoftNN loss stole a copy that achieves 76.9\% accuracy in 9,000 queries.







We discuss 5 current defenses against model extraction, including Prediction Poisoning~\citep{orekondy2019prediction}, Extraction Detection~\citep{juuti2019prada}, Watermarking~\citep{jia2020entangled}, Dataset Inference~\citep{maini2021dataset}, and Proof-of-Work~\citep{powDefense}.
We show that for each of these defenses, the existing implementations are inadequate in the SSL setting, and that retrofitting them to the specificities of encoders is non-trivial.
We identify two main sources of this difficulty: the lack of labels and the higher information leakage from the model outputs.



Our main contributions are as follows:
\begin{itemize}
    \item We call attention to a new setting for model extraction: the extraction of encoders trained with SSL. We find that the high dimensionality of their outputs make encoders particularly vulnerable to such attacks.
    \item We introduce and compare several novel model extraction attacks against SSL encoders and show that they are a realistic threat. In some cases, an attacker can steal a victim model using less than $\nicefrac{1}{5}$ of the queries required to train the victim.
    \item We discuss the effectiveness of existing defenses against model extraction in this new setting, and show that they are either inadequate or cannot be easily adapted for encoders.
    The difficulty stems from the lack of labels, which are utilized by some  defenses, and the higher information leakage in the outputs compared to supervised models.
    We propose these challenges as avenues for future research.
\end{itemize}

\section{Background and Related Work}

\subsection{Self-Supervised Learning}

Self-Supervised Learning (SSL) is an ML paradigm that trains models to learn generic representations from unlabeled data.
The representations are then used by downstream models to solve specific tasks.
The training signal for SSL tasks is derived from an implicit structure in the input data, rather than explicit labels.
For example, some tasks include predicting the rotation of an image~\citep{gidaris2018unsupervised}, or predicting masked words in a sentence~\citep{devlin2018bert}.
These tasks teach models to identify key semantics of their input data and learn representations that are generally useful to many downstream tasks due to their similarity structure.
Contrastive learning is a common approach to SSL where one trains representations so that similar pairs have representations close to each other while dissimilar ones are far apart.

\subsection{Self-Supervised Learning Frameworks}
Many new SSL frameworks have been proposed over the recent years; for ease of exposition, we focus on three current popular methods in the vision domain.

The SimCLR~\citep{simclr} framework learns representations by maximizing agreement between differently augmented views of the same sample via a contrastive InfoNCE loss~\cite{cpc} in the latent space.
The architecture is a Siamese network that contains two encoders with shared weights followed by a projection head.
During training, the encoders are fed either two augmentations of the same input (positive pair) or augmentations of different inputs (negative pair).
The output representations of the encoders are then fed into the projection head, whose outputs are used to compute a distance in the latent space; positive pairs should be close, and negative pairs far.
The encoders and the projection head are trained concurrently.
The authors show that the crucial data augmentation to create positive samples and achieve strong performance is a combination of a random resized crop and color distortion.


SimSiam~\citep{SimSiam} is another Siamese network architecture that utilizes contrastive learning like SimCLR but simplifies the training process and architecture.
In SimCLR, negative samples are needed during training to prevent the model from collapsing to a trivial solution, i.e. outputting the same representation for all inputs.
The authors of SimSiam show that negative samples are not necessary; collapse can be avoided by applying the projection head to only one of the encoders (in alternation) and preventing the gradient from propagating through the other encoder during training.

Barlow Twins~\citep{zbontar2021barlow} aims to create similar representations of two augmentations of the same image while reducing redundancy between the output units of the representations. It measures the cross-correlation matrix between two views and optimizes the matrix to be close to the identity matrix. The optimization process uses a unique loss function involving an invariance term and a redundancy reduction term. 
As with other methods, a projection head is used before passing the representation to the loss function.

\subsection{Model Extraction Attacks}
In the model extraction attacks for SL, an attacker queries the victim model to label its training points~\citep{pred_apis}. The main goals of such an adversary are to achieve a certain task accuracy of the stolen model~\citep{orekondy2019knockoff} or recreate a high fidelity copy that can be used to mount other attacks~\citep{fidelity}, such as the construction of adversarial examples~\citep{szegedy2013intriguing,papernot2017practical}. The attacker wants to minimize the number of queries required to achieve a given goal.
In the self-supervised setting, the goal of an adversary is to learn high-quality embeddings that can be used to achieve high performance on many downstream tasks.

\ahmad{Nicolas previously said to merge this section with the intro. Are we going to do that or is at fine as is?}
\jonas{given that we have the space, I'd say we can keep this here for now. It provides a bit more exposition/context, and the intro is pretty long already so I don't want to pack more in. But if we need space this is at the top of the cut list}

\subsection{Defenses against Model Extraction}
Defenses against model extraction can be categorized based on when they are applied in the extraction process.
They can be classified as proactive, active, passive, or reactive.

Proactive defenses prevent the model stealing before it happens. One of the methods is based on the concept of proof-of-work~\citep{powDefense}, where the model API users have to expand some compute by solving a puzzle before they can read the desired model output. The difficulty of the puzzle is calibrated based on the deviation of a user from the expected behavior for a legitimate user. 

Active defenses introduce small perturbations into model outputs to poison the training objective of an attacker~\citep{orekondy2019prediction} or truncate outputs~\citep{pred_apis}, however, these active methods lower the quality of results for legitimate users.

Passive defenses try to detect an attack~\citep{juuti2019prada}. Reactive defenses are post-hoc methods that are used to determine if a suspect model was stolen or not. The examples of such methods are watermarking~\citep{jia2020entangled}, dataset inference~\citep{maini2021dataset}, and Proof-of-Learning (PoL)~\citep{jia2021proof}. The PoL method can be immediately applied to SSL since defenders can claim ownership of a model by showing proof of incremental updates from their SSL model training.
\section{Extraction Methods}

\begin{figure}[t]
\vskip 0in
\begin{center}
\centerline{\includegraphics[width=\columnwidth]{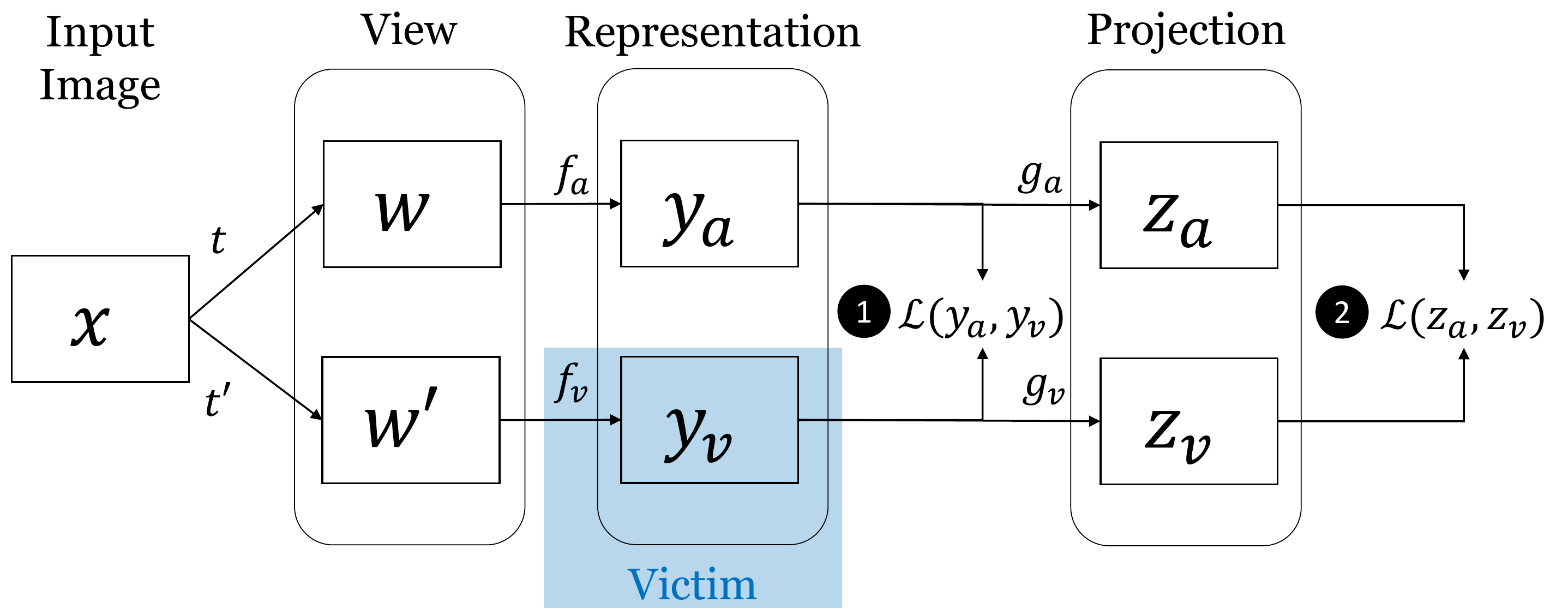}}
\caption{\textbf{Stealing the encoder} model $f_v$ using \encircle{1} the direct comparison between victim and attacker's representations or \encircle{2} access to the latent vectors $z$ or recreating the projection head $g_v$.}
\label{fig:byol-steal}
\end{center}
\vskip -0.3in
\end{figure}

\subsection{Threat Model}
The adversary makes queries to the victim model for representations of arbitrary inputs. The adversary has direct access only to the output representations. We consider settings where the adversary has/does not have knowledge about the encoder architecture, the victim's data augmentation scheme, loss function used, victim's training/test set or external dataset, and other parameters and hyper-parameters set by the victim during training of the encoder model.

\subsection{Extraction Algorithms}
\label{sec:algorithms}

We consider different approaches to extracting encoders trained with SSL. The final outcome of the self-supervised training process is an encoder or feature extractor $f$, where for an input $x$ we obtain the representation $y=f(x)$. The goal of an adversary is to steal the encoder $f$. Figure~\ref{fig:byol-steal} presents the stealing process from the attacker's perspective. An input image $x$ is augmented using two separate operators $t$ and $t'$ to obtain two correlated views
$w=t(x)$ and $w'=t'(x)$. Here a view refers to a version of the image under a particular augmentation. Note that it is possible that $t(x) = x$ in the case when no augmentations are used and that $w = w'$ when the same augmentation operator is used by the victim and attacker. 
The view $w$ is passed through the attacker's encoder $f_a$ to obtain  representation $y_a$ and $w'$ is fed to the victim's encoder $f_v$ which returns $y_v$. The representations can be further passed to the victim's head $g_v$ or attacker's $g_a$ to obtain latent vectors $z_v$ or $z_a$, respectively. To train the stolen encoder, our adversary adopts the methodology of  Siamese networks, for example, SimSiam~\citep{SimSiam}, where one of the branches is the victim's encoder and the other is the attacker's stolen local model. The attacker tries to train the stolen model to output representations as similar as possible to the victim encoder.

\paragraph{Direct Extraction.} The first plain attack method directly steals the encoder $f$ by comparing the victim representation $y_v$ with the representation $y_a$ obtained on the attacker's end through an appropriate loss function such as Mean Squared Error (MSE) or InfoNCE~\cite{cpc}. Algorithm~\ref{alg:stealing} presents the pseudo-code for stealing the representation model using the direct extraction method. 

\begin{algorithm}[t]
  \caption{Stealing an Encoder.}\label{alg:stealing}
  
  \algorithmicrequire{Querying Dataset $\mathcal{D}$, access to a victim model $f_v(w; \theta_v)$.}
  
  \algorithmicensure{Stolen representation model $f_a(w; \theta_a)$} 

  \begin{algorithmic}[1]
  \STATE Initialize $f_a$ with a similar architecture as $f_v$.
  \FOR{sampled queries $\{x_k\}_{k=1}^{N} \in \mathcal{D}$}
    \STATE Sample augmentations $t$ and $t'$.
    \STATE Generate views $w_k=t(x_k)$ and $w_k'=t'(x_k)$.
    \STATE Query victim model to obtain representations:\newline $y_{v} = f_v(w_{k}')$ 
    \STATE Generate representations from stolen model: \newline $y_{a} = f_a(w_{k}) $  
    \STATE Compute loss $\mathcal{L} \left\{ y_v, y_a \right\}$. 
    \STATE Update stolen model parameters $\theta_a \coloneqq \theta_a - \eta \nabla_{\theta_a} \mathcal{L}$.
    \ENDFOR
    
  \end{algorithmic}
\end{algorithm}

  
  

    

\textbf{Recreate Projection Head.}
The projection head $g$ is a critical component in the training process used for many contrastive learning approaches as shown by~\citet{simclr}. However, it is usually discarded after the training process.
Another method of stealing a model is thus to first recreate the victim's head and then pass the representations obtained from the victim through the head before finding the loss (see the pseudo-code in Algorithm~\ref{alg:recreate-head}). In practice, recreating the head requires knowledge of the victim's training process such as the augmentations and the loss function used which may not be possible in all cases. 
At the end of the attack spectrum, we also assess the case where the victim releases the projected representations $z$ or an adversary can access the head $g_v$.

Our adopted methodology is the \textit{Direct Extraction} method since, as we show in the evaluation, this method is able to outperform or match the performance of more sophisticated methods such as the head recreation. Moreover, compared to these more sophisticated methods, which require the use of augmentations, direct extraction is less prone to being detected by a defender (as we discuss in Section~\ref{sec:detection-defense}). 

\begin{algorithm}[t!]
  \caption{Recreating a Projection Head.}
  \label{alg:recreate-head}
  
  \algorithmicrequire{Querying dataset $\mathcal{D}$, access to a victim model $f_v(w; \theta_v)$.}
  
  \algorithmicensure{Recreated victim head $g_v(y; \theta_v')$ as $g_a(y; \theta_a')$.}

  \begin{algorithmic}[1]
  \FOR{sampled queries $\{x_k\}_{k=1}^{N} \in \mathcal{D}$}
    \STATE Sample augmentations $t$ and $t'$.
    \STATE Generate views $w_k=t(x_k)$ and $w_k'=t'(x_k)$.
    \STATE Query victim model to obtain representations: \newline $y_k = f_v(w_k)$ and $y_k' = f_v(w_k')$.
    \STATE Pass representations through head:  $z_k = g_a(y_k)$ and $z_k' = g_a(y_k')$.
    \STATE Compute loss $\mathcal{L} \left\{ z_k, z_k' \right\}$. 
    \STATE Update parameters of the head $\theta_a' \coloneqq \theta_a' - \eta \nabla_{\theta_a'} \mathcal{L}$.
    \ENDFOR
  \end{algorithmic}
\end{algorithm}
\subsection{Loss Functions}
\label{sec:loss-functions}

	

One of the most important hyper-parameter choices for representation stealing is the loss function, which can be categorized into quadratic, symmetrized, contrastive, and supervised-contrastive. They can be arranged into a hierarchy as in Figure~\ref{fig:loss-hierarchy}. The standard \textit{quadratic loss} like MSE (Mean Squared Error) can be used to directly compare the representations from the victim and the stolen copy. For example, we train a stolen copy of the model with the MSE objective such that it minimizes the $\ell_2$-distance between its predicted representations and the representation outputs from the victim model. 

The \textit{symmetrized losses}~\citep{byol,SimSiam}, have two parts to the loss
for a given data point $x$. When considering two separate views of $x$, $w$ and $w'$, the first part of the loss is $\mathcal{L}_1(g(f(w)),f(w'))$, where $f$ is the feature extractor, and $g$ represents the prediction head, which transforms the output of one view and matches it to the other view. More simply, the head $g$ is a small fully-connected neural network which transforms the representations from one space to another. 
The symmetry is achieved by computing the other part of the loss $\mathcal{L}_2(g(f(w')),f(w))$ and the final symmetrized loss is $\mathcal{L}=\mathcal{L}_1 + \mathcal{L}_2$. The loss functions $\mathcal{L}_1$ and $\mathcal{L}_2$ are commonly chosen as to be the negative cosine similarity. The standard supervised and symmetrized losses take into account only the distances between the representations and compare solely positive examples, i.e., the representations for the same input image. 

Finally, modern batch \textit{contrastive approaches}, such as InfoNCE~\citep{cpc} or Soft Nearest Neighbor (SoftNN)~\citep{softNearestNeighborLoss} compare not only positive but also the negative pair samples in terms of their distances and learn representations so that positive pairs have similar representations while negative pairs have representations which are far apart. The supervised contrastive loss (SupCon loss)~\citep{supconloss} is a novel extension to contrastive loss functions which uses labels from supervised learning in addition to positive and negative labeling. An attacker can use a labeled public dataset, such as ImageNet or Pascal VOC, to obtain the labels to be used with the SupCon loss. 



\section{Empirical Evaluation}
\label{experiments}

\subsection{Experimental Setup}

We include results for victim models trained on the ImageNet, CIFAR10, and SVHN datatsets. 
The ImageNet encoder has an output representation dimension of 2048, while encoders trained on CIFAR10 and SVHN return 512 dimensional representations. For ImageNet, we use the publicly available ResNet50 model from~\citep{SimSiam}. For the CIFAR10 and SVHN datasets, we use a public PyTorch implementation of SimCLR~\citep{simclr} to train victim ResNet18 and ResNet34 models over 200 epochs with a batch size of 256 and a learning rate of 0.0003 with the Cosine Annealing Scheduler and the Adam optimizer. For training stolen models, we use similar (hyper-)parameters to the training of the victim models. More details on the experimental setup are in Section~\ref{sec:exeperimental-setup-details}.

\subsection{Linear Evaluation}

The usefulness of representations is determined by evaluating a linear layer trained from representations to predictions on a specific downstream task. 
We follow the same evaluation protocol as in \citet{simclr}, where a linear prediction layer is added to the encoder and is fine-tuned with the full labeled training set from the downstream task while the rest of the layers of the network are frozen. The test accuracy is then used as the performance metric. 

\subsection{Methods of Model Extraction}

We compare the stealing algorithms in Table~\ref{tab:compare-attack-methods}. The \textit{Direct Extraction} steals $f_v$ by directly comparing the victim's and attacker's representations using the SoftNN loss, the \textit{Recreate Head} uses the SimSiam method of training to recreate the victim's head which is then used for training a stolen copy, while \textit{Access Head} trains a stolen copy using the latent vectors $z_v$ and the InfoNCE loss. For CIFAR10 and STL10 datasets, the \textit{Direct Extraction} works best, while it is outperformed on the SVHN dataset by the \textit{Access Head} method. 
The results for the \textit{Recreate Head} 
are mixed; the downstream accuracy for loss functions such as InfoNCE improves when the head is first recreated while for other loss functions such as MSE, 
which directly compares the representations, the recreation of the head hurts the performance. This is in line with the SSL frameworks that use the InfoNCE loss on outputs from the heads~\citep{simclr}.
The results show that the representations can be stolen directly from the victim's outputs; the recreation or access to the head is not critical for downstream performance.

We observe that bigger differences in performance stem from the selection of the loss function---a comparison of the different loss functions is found in Table~\ref{tab:compare-losses}. The choice of the loss function is an important parameter for stealing with the SoftNN stolen model having the highest downstream accuracy in two of the three datasets while InfoNCE has a higher performance in one of the tasks.  However, as we show in Table~\ref{tab:combinednumqeury}, the number of queries used is also a factor behind the performance of different loss functions.  We note that the SupCon loss gives the best performance on the CIFAR10 downstream task and this is likely a result of the fact that SupCon assumes access to labeled query data, which in this case was the CIFAR10 test set.


The results in Table~\ref{tab:StealImagenet} show the stealing of the encoder pre-trained on ImageNet. Most methods perform quite well in extracting a representation model at a fraction of the cost with a small number of queries (less than one fifth) required to train the victim model as well as augmentations not being required. In general, the \textit{Direct Extraction} with InfoNCE loss performs the best and the performance of the stolen encoder increases with more queries.

\begin{table}[t]
\caption{\textbf{Comparison between attack methods} for the classification top-1 accuracy on downstream tasks CIFAR10, STL10, and SVHN. The models are stolen from a CIFAR10 victim encoder with 9,000 queries from the CIFAR10 test set.
}
\label{tab:compare-attack-methods}
\vskip 0.15in
\begin{center}
\begin{small}
\begin{sc}
\begin{tabular}{lcccr}
\toprule
Method\textbackslash Dataset & CIFAR10 & STL10 & SVHN \\
\midrule
\textit{Victim Model} & 79.0 & 67.9 & 65.1 \\
\cdashlinelr{1-4}
\textbf{Direct Extraction} & \textbf{76.9} & \textbf{67.1} & 67.3 \\
\textbf{Recreate Head} & 59.9 & 52.8 & 56.3  \\
\textbf{Access Head}  & 75.6 & 65.0 & \textbf{69.8}  \\
\bottomrule
\end{tabular}
\end{sc}
\end{small}
\end{center}
\vskip -0.1in
\end{table}
\begin{table}[t]
\caption{\textbf{Comparison between loss functions.} We use the same setup as in Table~\ref{tab:compare-attack-methods}. Loss functions with (*) use data augmentations.}
\label{tab:compare-losses}
\vskip 0.15in
\begin{center}
\begin{small}
\begin{sc}
\begin{tabular}{lccc}
\toprule
Loss Type\textbackslash Dataset & CIFAR10 & STL10 & SVHN \\
\midrule
\textit{Victim Model} & 79.0 & 67.9 & 65.1 \\
\cdashlinelr{1-4}
MSE    & 75.5 & 64.8 & 58.8 \\
InfoNCE & 75.5 & 64.6 & \textbf{69.6}  \\
SoftNN     & \textbf{76.9} & \textbf{67.1} & 67.3 \\
Wasserstein    & 63.9 & 50.8 & 49.5 \\
\cdashlinelr{1-4}
SupCon*      & \textbf{78.5} & 63.1 & 67.1 \\
Barlow*     & 26.9 & 26.6 & 26.2 \\
\bottomrule
\end{tabular}
\end{sc}
\end{small}
\end{center}
\vskip -0.1in
\end{table}

\begin{table*}[t]
\caption{\textbf{Linear Evaluation Accuracy} on a victim and stolen encoders. The victim encoder is pre-trained on the ImageNet dataset. 
}
\label{tab:StealImagenet}
\begin{center}
\begin{small}
\begin{sc}
\begin{tabular}{ccccccccc}
\toprule
Loss & \# of Queries & Dataset & Data Type & CIFAR10 & CIFAR100 & STL10 & SVHN & F-MNIST\\
\midrule
\textit{Victim Model} & N/A & N/A & N/A & 90.33 & 71.45 & 94.9 & 79.39 & 91.9 \\
\cdashlinelr{1-9}
MSE & 50K & CIFAR10 & Train & 75.7 & 43.9 & 27.3 & 48.7 & 61.7  \\
InfoNCE & 50K & CIFAR10 & Train & 61.8 & 32.9 & 56.8 & 51.5 & 82.1 \\
MSE & 50K & ImageNet & Test & 47.7 & 19.3 & 13.1 & 23.8 & 82.1 \\
MSE & 50K & ImageNet & Train & 51.0 & 17.2 & 49.5 & 39.7 & 84.7 \\ 
InfoNCE & 50K & ImageNet & Test & 64.0 & 35.3 & 60.4 & 63.7 & 88.7 \\
InfoNCE & 50K & ImageNet & Train & 65.2 & 35.1 & 64.9 & 62.1 & 88.5 \\ 
MSE & 100K & ImageNet & Train & 55.5 & 23.0 & 27.1 & 27.3 & 82.1 \\ 
InfoNCE & 100K & ImageNet & Train & 68.1 & 38.9 & 63.1 & 61.5 & 89.0 \\
MSE & 200K & ImageNet & Train & 56.1 & 22.6 & 47.8 & 55.4 & 87.6 \\
InfoNCE & 200K & ImageNet & Train & 79.0 & 53.4 & 81.2 & 48.3 & \textbf{90.5} \\
InfoNCE & 250K & ImageNet & Train & \textbf{80.0} & \textbf{57.0} & \textbf{85.8} & \textbf{71.5} & 90.2 \\
\bottomrule
\end{tabular}
\end{sc}
\end{small}
\end{center}
\vskip -0.1in
\end{table*}



\subsection{Comparing Stealing Encoders vs Stealing Supervised Models}

To test the hypothesis that the higher information leakage from encoders makes them more vulnerable to extraction attacks than supervised models, we compare stealing encoders (using representations) against stealing supervised models (using labels). Here we observe a higher accuracy on downstream tasks after stealing with representations compared to labels. For example, we steal a SimCLR victim model pretrained on the CIFAR10 train set using both labels and embeddings. The accuracy of the stolen model is 75.5\% when using representations whereas it is 66.9\% when stealing with labels. To use labels, the SimCLR encoder with a finetuned final layer (the downstream classifier on CIFAR10) is used as the victim model. 
Full results for these two settings are shown in Table~\ref{tab:steal_ssl_vs_sl}. 
\textit{Fine-tune} in the case of SL means that given the stolen label model, we again fine tune it on the downstream task by removing the final layer and repeating the linear evaluation process which is run when stealing encoder models. \textit{No fine-tune} refers to the accuracies directly obtained from the stolen model. 


\begin{table}[t]
\caption{\textbf{Stealing embeddings vs labels.} We compare SSL (Self-Supervised Learning) methods with embeddings vs SL (Supervised Learning) methods with labels, posteriors, or logits. The victim models are (pre)trained on CIFAR10 train set.}
\label{tab:steal_ssl_vs_sl}
\vskip 0.15in
\small
    \centering
        \begin{tabular}{lccc}
        \toprule
        \textbf{Method} 
        & \textbf{ CIFAR10 } &  \textbf{ STL10 } &  \textbf{ SVHN } \\
        \midrule
        \textit{Victim SSL} 
        & 79.0 & 67.9 & 65.1  \\
        \cdashlinelr{1-4} 
        SSL stealing 
        & 75.5 & 64.6 & 69.6 \\
        SL no fine-tune 
        & 65.6 & 52.4 & 32.1 \\
        SL fine-tune 
        & 66.9 & 56.0 & 38.1 \\
        \bottomrule
        \end{tabular}
\end{table}
\section{Defense Strategies}
We consider different defense strategies to protect representation models from model extraction attacks. We analyze current state-of-the-art strategies from SL and either show why they are inadequate for SSL in their present state or how they can be adjusted to the new setting.



\subsection{Watermarking-based Defense}
\label{sec:watermarking-defense}

SL methods have been shown to use input triggers and labels to watermark their models and claim the ownership with high confidence~\cite{jia2020entangled}. In the self-supervised setting, labels are replaced by vector representations. Adding a trigger to the input may not entangle the watermark enough to be extracted during stealing, and adding a trigger to the representations leaves the defense susceptible to adaptive attacks like pruning. Instead, we can take advantage of data augmentations to incorporate structural watermarks in representations that are naturally extracted during stealing. In particular, we select a private augmentation, not used during contrastive training. We train representations to contain discriminative information about this augmentation instead of being invariant to it. We then use the presence of this information to claim ownership, without affecting downstream performance. 

Such a defense requires training the encoder simultaneously with an \augpredict. This is a relatively small discriminator and adds a negligible computational cost to training. It is trained to distinguish between views of the private augmentation and the encoder is trained to output representations that perform well on the \augpredict, in addition to the contrastive learning. Representations from this encoder are invariant to other augmentations but contain discriminative information about the private one. We hypothesize that any stolen model would extract this information and also perform well on the \augpredict. On the other hand, other genuinely trained encoders should perform no better than random chance on it.
 
 
We test our hypothesis on the \textit{Direct Extraction} attack against SimCLR encoders for CIFAR10 and SVHN datasets. The original SimCLR training procedure uses random crops, horizontal flips, color jitter, grayscale and Gaussian blur for contrastive learning of representations. We use rotation prediction as the watermarking task. For every input image, we randomly generate two views, one rotated by an angle in $[0^{\circ}, 180^{\circ}]$ and the other rotated by an angle in $[180^{\circ}, 360^{\circ}]$. The victim encoder includes a two-layer MLP, the \augpredict, that maps representations to the rotation interval they belong to. The encoder optimizes both the training and watermarking objectives. When evaluating Watermark Success Rate, training images are selected and augmented with the same rotations the victim is trained with. The stolen model's representations are then evaluated on the \augpredict to obtain its accuracy on the watermark task. Note that the same evaluation of another genuinely trained model gives an accuracy of around 50\%, close to random chance. 

\begin{figure}[t]
    \centering
    \includegraphics[width=1.0\linewidth]{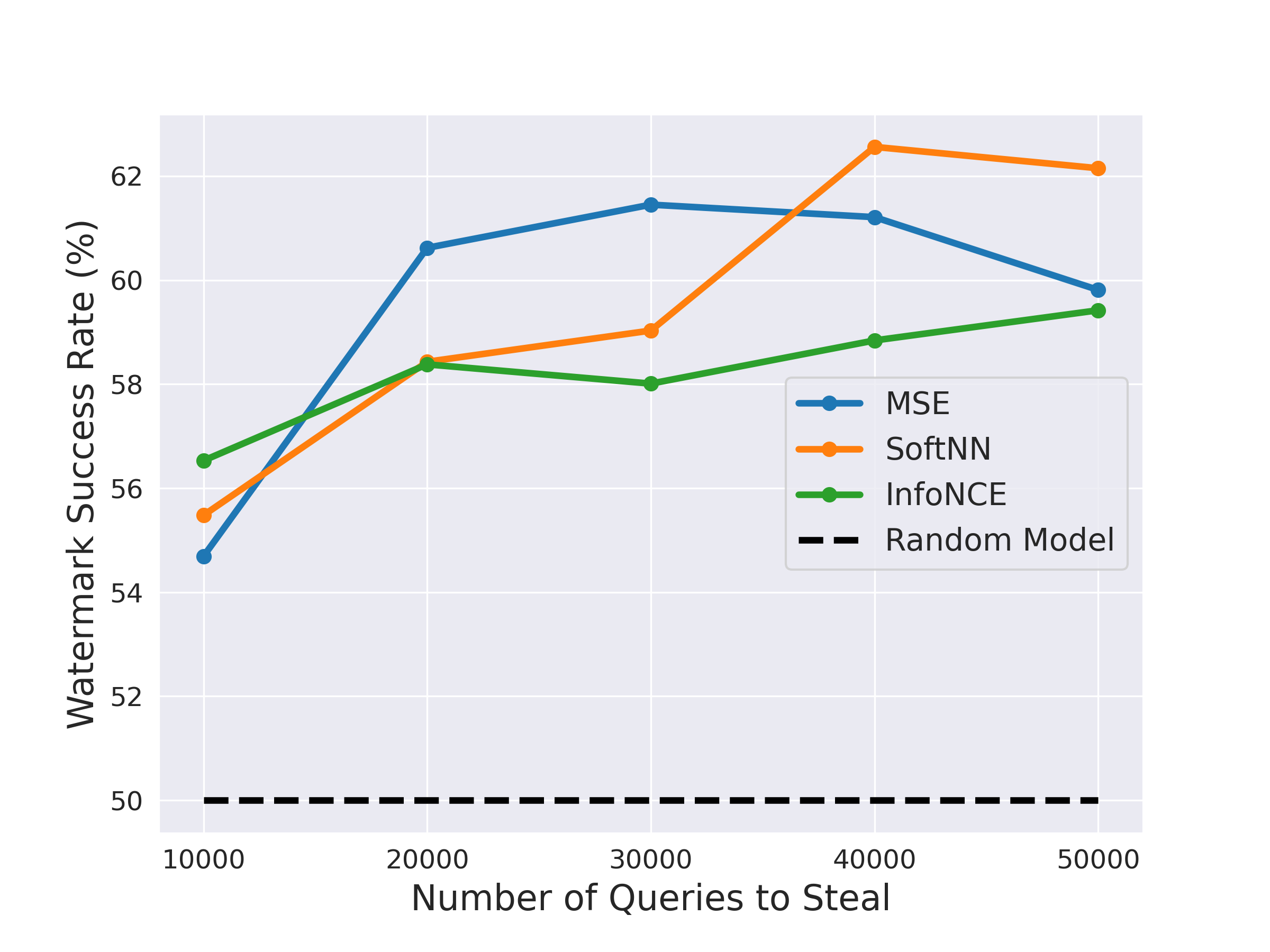}
    \vskip -0.1in
    \caption{\textbf{Watermark Success Rate} for models stolen with different loss functions vs. a genuinely trained model (Random Model). 
    }
    \label{fig:wm_graph}
    \vskip -0.1in
\end{figure}

To claim the ownership of the stolen model, we use the statistical t-test. The null hypothesis is that the probability of correct rotation classification for a tested model should be equal to those of a benign random model. Our results in Table~\ref{tab:ownership-watermark} and Figure~\ref{fig:wm_graph} give evidence that the watermarked behavior transfers to models stolen using different loss functions. These models obtain accuracies on the watermark task that are significantly higher than random chance and can be used to claim the ownership with 95\% confidence. While this method inherently entangles watermark behavior in the representations, an adaptive adversary may attempt to remove it. We note that such an adaptive attack is unlikely to succeed due to the absence of input-label pairs that are key to SL methods. As such, finding the private augmentation and stealing the victim while enforcing invariance would require as much insight and computational cost as training from scratch. However, we leave the complete treatment of this to future work.


We note the assumptions, advantages and disadvantages of this method. The success of this defense hinges on two heavy assumptions: (1) genuinely trained models would not reasonably output representations that contain this exact discriminative information. In practice, they may actually be invariant to it. (2) The adversary is unable to guess the watermark task and remove its information without essentially training from scratch. If these assumptions hold, this defense has the advantage of inherently entangling the watermark task information into the representation. Even a model stolen with fewer queries extracts the watermark well enough for the defender to claim ownership, as demonstrated in Figure~\ref{fig:wm_graph}. Disadvantages include failure modes when the above assumptions do not hold, and the requirement to retrain existing encoders with an \augpredict. This may be difficult especially if a model has already been trained.

\begin{table}[t]
\caption{\textbf{Ownership t-test for Watermarks.}  $\Delta \mu$ is the effect size.}
\label{tab:ownership-watermark}
\vskip 0.15in
\begin{center}
\begin{small}
\begin{sc}
\begin{tabular}{ccccc}
\toprule
Dataset & \# Queries & $\Delta \mu$ & p-value & t-value \\
\midrule
CIFAR10 & 10K & 0.07 & 6.34e-14 & 9.81 \\
CIFAR10 & 50K & 0.11 & 1.27e-21 & 15 \\
SVHN & 10K & 0.06 & 2.59e-11 & 8.22 \\
SVHN & 50K & 0.09 & 3e-15 & 10.64 \\
\bottomrule
\end{tabular}
\end{sc}
\end{small}
\end{center}
\vskip -0.1in
\end{table}

\subsection{Perturbation-based Defenses}

The perturbation-based defenses modify model outputs to hinder the training of the stolen copy. A simple application of this method for SSL is to perturb the output representations with some random noise. Due to the large dimensions and different use cases of representations compared to softmax outputs in the supervised setting, the results of adding noise highly depends on the training objectives of an attacker and legitimate user. 
We conducted a simple test of this adaptation and observed a drop of up to 10\% in the performance of a stolen encoder against a non-adaptive attacker, while also incurring a slight drop in performance (of around 1\% to 2\%) on downstream tasks for legitimate users. (see Section~\ref{sec:add-noise-to-representations} in Appendix).
However, in general, this noise needs to be tuned according to the situation and may require further assumptions about attackers. 

Prediction poisoning is an advanced active defense method against the extraction of SL models, where the defender perturbs the outputs to decrease the performance of the attacker's stolen model~\citep{orekondy2019prediction} while maintaining a high quality of outputs for benign users.
In the case of a prediction poisoning defense for SSL, the defender should add perturbations to the outputs that must simultaneously optimize two objectives: the perturbation should harm the training of an encoder model by an adversary, while not affecting the training of a downstream model used by a legitimate user. We provide an analysis that shows the difficulty in applying this defense in the SSL setting (see Section~\ref{sec:prediction-poisoning-analysis}).

The prediction poisoning defense for SSL requires a complicated and expensive two-way optimization. The victim would have to use at least two surrogate models, for the stolen encoder and a legitimate downstream classifier, and approximate their respective Jacobians. Moreover, several assumptions are needed such as the type of loss functions used and the types of potential downstream tasks legitimate users are training on. Another issue is the lack of labels in SSL. The victim must know the target labels $L$ used by the legitimate user for the specific downstream task. This can be done by considering multiple possibilities for downstream tasks or by requiring legitimate users to choose a specific downstream task. However, since there can be multiple legitimate users, multiple possible values of $L$ would need to be considered, thereby further constraining the optimization.



\subsection{Detection of Model Extraction}
\label{sec:detection-defense}
The passive defenses try to detect adversarial inputs by analyzing the distribution of input queries. However, there is a very wide spectrum of acceptable inputs to SSL models. Moreover, detection methods like PRADA~\citep{juuti2019prada} do not generalize to attacks that utilize only natural data~\citep{orekondy2019knockoff,pal2020activethief} while such data can be used for the representation extraction. The detection methods might be useful against attackers who apply contrastive loss functions (e.g., InfoNCE or SupCon) and provide two augmentations of the same image as inputs, for example, to recreate the head $g$. For this purpose, the victim can collect the history of representations returned for a given user. Then, for a new query $w$, it can check the history to determine if there is a similar query $w'$, potentially different augmentation of the same raw input image $x$.

Most encoders
are trained to maximize the similarity on outputs from a projection head $z=g(\cdot)$.
Thus, we carry out the comparison between queries on the level of their projection outputs $z$ from the head $g_v$ instead of on representations $y$. We do not carry out the similarity search on the level of input images since the representations or projection outputs can be seen as compressed versions of inputs and stored at a much lower cost.
There are different metrics that can be used to measure the distances between vectors. We find that both the $\ell_2$-norm and cosine distance metrics work relatively well in identifying similar queries (Tables~\ref{tab:compare-threshold},\ref{tab:compare-threshold3}). 
Moreover, using the head is important in improving the performance as we observe lower error rates (Tables~\ref{tab:compare-threshold},\ref{tab:compare-threshold2}). 
If queries for different views of the same input are run from different accounts, then the defense has to analyze representations from many accounts and find out which accounts issue queries that result in similar representations. 



\subsection{Dataset Inference}

Another defense that we consider is inspired by the idea of dataset inference~\citep{maini2021dataset}. It identifies whether a given model was stolen by verifying if a suspected adversary's model has private knowledge from the victim's train set. It builds on the observation that a model behaves differently when presented with its training data versus unseen (test) data; the private training points are further away from decision boundaries than public (test) data points since the optimization algorithms locally \textit{overfit} around the decision boundaries.
Then, the stolen model, although obtained by model extraction, exhibits similar behavior to the victim model.
The self-supervised models do not have decision boundaries, so we adjust the dataset inference defense: instead of analyzing distances of data points from decision boundaries, we analyze loss values from private (train) vs public (test) sets. To compute the loss, we use the projection head from the training stage. \ahmad{This is still somewhat too focused on the results, highlight the difficulties}


\textbf{Hypothesis Testing.} Using the loss scores produced by head $g_v$, we create equal-sized sample vectors $l_t$ and $l_p$ from private training and public data, respectively. Following \cite{maini2021dataset}'s method, we test the null hypothesis $\mathcal{H}_0 : \mu_p < \mu_t$ where $\mu_p=\bar{l_p}$ and $\mu_t = \bar{l_t}$ are mean loss scores. The test
either rejects $\mathcal{H}_0$ and conclusively rule that $f_v$ is \textit{stolen}, or gives an inconclusive result.

Our results suggest that the signal for identifying that a stolen model contains knowledge about a victim's private training data is much weaker for encoders than SL models.
We hypothesize that SL models \textit{overfit} more to the training set than the encoders, which apply heavier augmentations (see Table~\ref{tab:dataset-inference-main}). This makes it harder to use dataset inference in the SSL setting.

\subsection{Calibrated Proof-of-Work}
The calibrated proof-of-work defense proposes to make model extraction more difficult by requiring users to solve a puzzle before they can read predictions from an ML model exposed via a public API~\citep{powDefense}. The puzzle is generated using the HashCash function and requires a user to find a suffix that can be appended to a challenge string so that the message digest has a specific number of leading zeros. The difficulty of the puzzle depends on how the privacy leakage incurred by a given user differs from the expected privacy leakage for a legitimate user.

The defense requires us to model a behavior of a legitimate user. In the SL setting, we can expect queries to come from a similar distribution as the victim's training set. However, for the SSL models, the training set is usually much larger and more diverse. In principle, any user query might be valid since we cannot anticipate any specific downstream task. Moreover, this defense method estimates the information or privacy leakage from the target model. These can be easily computed based on the softmax outputs in SL but not directly from the representations, which are not trained to express probability values. 
The defense uses the PATE framework~\citep{papernot2017semi,papernot2018scalable}, which measures privacy leakage for individual queries. PATE creates teacher models that are trained on the defender’s training data and uses predictions from each
teacher. Such predictions are not available in the encoder models.  
Another way to estimate the information leakage of a query is by computing the entropy value. Again, it can be calculated from the softmax values in SL but not from the representations. A possible way to circumvent the difficulty is by computing the entropy directly on the input image, however, this value is not tied to the model and does not reflect its information leakage. Overall, we could require a user to solve a small puzzle for each query to prevent the ML API from being flooded by requests, but it is challenging to precisely calibrate the difficulty of the puzzle.
\section{Conclusions and Future Work}
As MLaaS shifts away from prediction APIs and towards representation APIs, we analyze how the encoders trained with SSL  can be extracted when  exposed to the public. We find that the direct objective of imitating the victim's representations gives high performance on downstream tasks despite the attack requiring only a fraction (less than 15\% in certain cases) of the number of queries needed to train the stolen encoder in the first place. 
With limited queries, performance degrades rather quickly. We hypothesize this is due to a lack of useful inductive biases, \textit{e.g.}, data augmentations.
For the ImageNet encoder, the attacker's accuracy for downstream tasks is within 1\% of the victim's accuracy on simple tasks like Fashion MNIST, and up to 17\% for more complicated datasets like CIFAR100. 
The quality of the stolen encoders is highly dependent on the selection of the attacker's loss function and tends to be better with the MSE loss in the limited query regime while InfoNCE is more useful when more queries are available. 
However, it is challenging to defend encoders trained with SSL since the output representations leak a substantial amount of information. The most promising defenses are reactive methods, such as watermarking, that can embed specific augmentations in high-capacity encoders. Overcoming the discussed challenges and improving defenses is an important avenue for future work. Another interesting direction to extend the work on encoder extraction is in the natural language domain for large text representation models.

\section*{Acknowledgments}
We would like to acknowledge our sponsors, who support our research with financial and in-kind contributions:
CIFAR through the Canada CIFAR AI Chair program, DARPA through the GARD program, Intel, Meta, NFRF through an Exploration grant, and NSERC through the Discovery Grant and COHESA Strategic Alliance. Resources used in preparing this research were provided, in part, by the Province of Ontario, the Government of Canada through CIFAR, and companies sponsoring the Vector Institute. We would like to thank members of the CleverHans Lab for their feedback.


\bibliography{main}
\bibliographystyle{icml2022}

\newpage
\appendix
\onecolumn
\newpage

\onecolumn

\title{\papertitle (Supplement)}
\icmltitle{\papertitle\  (Supplement)}

\section{Additional Experimental Results}

\subsection{Details on Experimental Setup}
\label{sec:exeperimental-setup-details}

We test various attack methods with different loss functions for the stealing process. We use different numbers of queries from different datasets, where each query represents an input to the victim model. The user then receives as output the representation from the victim. 

For training a stolen model, we use similar (hyper-)parameters to the training of the victim models, with a batch size of either 64 or 256, initial learning rate of 0.0001, and the Adam optimizer. In the case of stealing from the ImageNet victim model, we use a larger learning rate of 0.1 or 1.0 with the LARS optimizer~\citep{ginsburg2018large} and a batch size of 256 or 512. 
    
For downstream tasks, we use the CIFAR10, SVHN, STL-10, and Fashion MNIST datasets to compare the performance of the stolen model to the victim model. We use the standard linear evaluation protocol where an additional linear layer is added to the representation model while all other layers are frozen. The network is then optimized with labeled data from the specific downstream task.  For the models stolen from a CIFAR10 or SVHN victim model, we use a learning rate of 0.0001 with the SGD optimizer for the linear evaluation with the parameters tuned for the victim model and then keep them constant while evaluating the models that are stolen from it. For the ImageNet victim model, we use a learning rate of 1.0 with the LARS optimizer and a batch size of 256. In all cases, the top 1 test accuracy on the specific downstream task was reported and used for the comparison between the victim and stolen models. 

We ran all experiments on machines equipped with an Intel\textregistered~Xeon\textregistered~Silver 4210 processor, 128 GB of RAM, and four NVIDIA GeForce RTX 2080 graphics cards, running Ubuntu 18.04.

\subsection{Perturbation based Defense Methods}

We evaluated several types of perturbation based defenses against extraction attacks involving representations. To model a legitimate user, we used a two layer linear network which is trained by the legitimate user to map from representations obtained from the victim model to the final label in the specific downstream task the user is interested in. Therefore once the network is trained, the legitimate user can obtain a label for a new image by first querying the victim model and then using the obtained representation as an input to his or her trained network to find the label. By contrast, an adversary will first attempt to steal the full model on the victim side by querying and then add an additional linear layer for a downstream task. The major difference between an adversary and legitimate user is that an adversary trains a full model which gives it the additional benefit of being able to be useful for various downstream task and not having to first query the victim model to get the label for a new input. Table~\ref{tab:compare-defence} and~\ref{tab:compare-defence2} shows the results of using simple Gaussian noise with a mean of 10 and 0 respectively and the accuracy the legitimate user and adversary can obtain for different levels of noise. The results show that increasing the level of noise added harms the adversary while the accuracy of the legitimate user remains relatively constant. Moreover, using a higher mean for the noise is more effective. Table~\ref{tab:compare-defencehigh} shows results for an alternative defence method which only perturbs representations it identifies as being close to a previously queried representation. The defence seeks to identify attacker's which use several augmentations of the same image under the assumption that the representations of two augmentations of the same image will be more similar than the representations of two different images. We use the $\ell_2$ distance and cosine distance to measure the similarity between two representations and set a threshold. In the case the distance between a new representation and a previously returned representation is under a set threshold, it is replaced by random Gaussian noise with a high mean and variance. From Table ~\ref{tab:compare-defencehigh}, we see that there is little effect of this defense on a legitimate user as desired while an adversary using multiple augmentations (such as with the InfoNCE loss) has a reduced accuracy. The standard deviation $\sigma$ of the noise however does not have a major impact on reducing the accuracy of the adversary apart from the initial drop from no noise to noise with $\sigma = 20$.

\subsection{Compare Loss Functions}

We further compare loss functions in Table~\ref{tab:compare-lossessvhn}.

\begin{table}[t]
\caption{\textbf{Comparison between loss functions} for the classification top-1 accuracy on downstream tasks and models stolen from a SVHN victim model. 9000 queries from the SVHN test set are used for stealing. Loss functions with (*) use data augmentations.}
\label{tab:compare-lossessvhn}
\vskip 0.15in
\begin{center}
\begin{small}
\begin{sc}
\begin{tabular}{lcccr}
\toprule
Loss\textbackslash Dataset & CIFAR10 & STL10 & SVHN \\
\midrule
Victim & 57.5 & 50.6 & 80.5  \\
\midrule
MSE    & 51.2 & 46.3 & 80.6 \\
Wasserstein    & 46.4 & 40.1 & 69.4  \\
InfoNCE & \textbf{56.3} & \textbf{50.4} & \textbf{86.2}  \\
SoftNN     & 48.4 & 44.6 & 78.6 \\
\midrule
SupCon*      & 42.3 & 33.9 & \textbf{92.1}  \\
SimSiam*      & 49.7 & 44.0 & 66.7 \\
Barlow*     & 17.9 & 16.3 & 19.4 \\
\bottomrule
\end{tabular}
\end{sc}
\end{small}
\end{center}
\vskip -0.1in
\end{table}

\begin{table}[t]
\caption{\textbf{Comparison between number of queries} from the attacker to the classification accuracies on downstream tasks of CIFAR10, STL10 and SVHN for the stolen representation model. The victim model was trained using the CIFAR10 dataset and the attacker used the \textbf{SVHN} training set for queries. The attacker used the \textbf{InfoNCE} loss.}
\label{tab:compare-queriessvhninfonce}
\vskip 0.15in
\begin{center}
\begin{small}
\begin{sc}
\begin{tabular}{lccccr}
\toprule
Queries\textbackslash Dataset & CIFAR10 & STL10 & SVHN  \\
\midrule
Victim Model & 79.0 & 67.9 & 65.1 \\
500    & 35.7 & 31.1  &  26.4 \\ 
1000     & 38.6 & 34.7 & 40.4  \\ 
5000     & 50.5 & 46.6 & 72.7  \\ 
10000   & 59.3 & 51.6 & 73.1  \\ 
20000    & 69.3 & 56.6 & 72.8  \\ 
30000      & 69.9 & 56.6 & 67.5  \\ 
50000      & 71.8 & 57.6 & 67.1  \\ 
\bottomrule
\end{tabular}
\end{sc}
\end{small}
\end{center}
\vskip -0.1in
\end{table}

\begin{table*}
\caption{\textbf{Comparison between number of queries} from the attacker to the classification accuracies on downstream tasks of CIFAR10, STL10 and SVHN for the stolen representation model. The victim model was trained using the CIFAR10 dataset and the attacker used the \textbf{SVHN} training set for queries. The attacker used the \textbf{SoftNN}, \textbf{MSE} and \textbf{InfoNCE} loss functions.}
\label{tab:combinednumqeury}
\vskip 0.15in
\begin{center}
\begin{small}
\begin{sc}
        \begin{tabular}{lcccccccccc}
        \toprule
                     \textbf{\# queries / Loss}  & & \textbf{SoftNN}         &  &  & \textbf{MSE}  & & & \textbf{InfoNCE}  \\
        \cmidrule(lr){1-1} \cmidrule(lr){2-4} \cmidrule(lr){5-7} \cmidrule(lr){8-10} 
        \textbf{SVHN} & \textbf{CIFAR10} & \textbf{STL10} & \textbf{SVHN} & \textbf{CIFAR10} & \textbf{STL10} & \textbf{SVHN} & \textbf{CIFAR10} & \textbf{STL10} & \textbf{SVHN} \\
        \cmidrule(lr){1-1} \cmidrule(lr){2-4} \cmidrule(lr){5-7}  \cmidrule(lr){8-10} 
        500 & 39.3 & 36.1 & 28.2  &  36.1  & 33.2 & 39.5  & 35.7 & 31.1 & 26.4 \\
        1000 & 40.0 & 39.6 & 40.9 &  43.4  &  39.7 & 58.1 & 38.6 & 34.7 & 40.4 \\
        5000 & 55.5  & 49.5 & 66.0  &  55.6  &  47.6   & 58.7 & 50.5 & 46.6 & 72.7 \\
        10000 & 60.4 & 53.4 & 65.9 &  59.3  &   50.0  &  58.0 & 59.3 & 51.6 & 73.1 \\
        20000 & 67.2  & 58.5 & 68.0 &  58.1  &  50.9   & 56.6 & 69.3 & 56.6 & 72.8 \\
        30000 & 70.8 &  58.5  & 67.3  &  60.6  &  51.0   & 58.7 & 69.9 & 56.6 & 67.5 \\
        50000 & 70.9 & 60.3 & 67.7 & 61.6 & 52.7 & 59.3 & 71.8 & 57.6 & 67.1 \\
        \midrule
        \textbf{Victim} & \textbf{79.0} & \textbf{67.9} & \textbf{65.1} & \textbf{79.0} & \textbf{67.9} & \textbf{65.1} & \textbf{79.0} & \textbf{67.9} & \textbf{65.1} \\
        \bottomrule
        \end{tabular}
    \end{sc}
\end{small}
\end{center}
\vskip -0.1in
\end{table*}

\begin{table}[t]
\caption{\textbf{Comparison between number of queries} from the attacker to the classification accuracies on downstream tasks of CIFAR10 and SVHN for the stolen representation model. The victim model was trained using the CIFAR10 dataset and the attacker used the \textbf{CIFAR10 training set} for queries.  The attacker used the \textbf{Soft Nearest Neighbours} (SoftNN) loss. \ahmad{Based on previous results. To remove.} }
\label{tab:compare-queries}
\vskip 0.15in
\begin{center}
\begin{small}
\begin{sc}
\begin{tabular}{lcccr}
\toprule
Queries\textbackslash Dataset & CIFAR10 & STL10 & SVHN  \\
\midrule
Victim Model & 67.3 & 74.5 & 59.3 \\
500    & 47.7 & 39.3 & 45.1 \\
1000    & 51.8 & 41.8 & 53.2 \\
5000    & 58.4 & 49.2 & 59.1 \\
10000 & 63.9 & 53.1 & 57.8 \\
20000 & 67.5 & 55.3 & 58.2   \\
30000     & 66.6 & 55.7 & 60.0 \\
40000      & 65.5 & 56.3 & 58.8 \\
50000      & 66.3 & 56.7 & 59.8 \\
\bottomrule
\end{tabular}
\end{sc}
\end{small}
\end{center}
\vskip -0.1in
\end{table}

\subsection{Adding Random Noise to Representations}
\label{sec:add-noise-to-representations}

\begin{table}[t]
\caption{\textbf{Noise based perturbations} as a defence. The downstream accuracy on CIFAR10 for a legitimate user and an adversary are shown where the adversary used the \textbf{MSE} loss. All results are based on 9000 queries from the CIFAR10 test set. The noise added was Gaussian noise with a mean of 10 and standard deviation of $\sigma$ as in the table. }
\label{tab:compare-defence}
\vskip 0.15in
\begin{center}
\begin{small}
\begin{sc}
\begin{tabular}{lccr}
\toprule
$\sigma$ \textbackslash Dataset & Legitimate & Adversary \\
\midrule
0 & 67.1 & 66.3 \\
1    & 65.7 & 56.9  \\
2  & 67.9 & 56.8 \\
3 & 64.0 & 56.9 \\
4    & 62.2 & 55.8\\
\bottomrule
\end{tabular}
\end{sc}
\end{small}
\end{center}
\vskip -0.1in
\end{table}

The first approach to perturbing the representations is to add random Gaussian noise as in Table~\ref{tab:compare-defence}. In general, the representations are resilient to noise and rather a large amount of noise has to be added to decrease the performance on downstream tasks. We observe that such perturbations can even slightly increase the accuracy on the downstream task since they can act as regularizers. Overall, when we add noise, for a legitimate user - we decrease the accuracy $A_L$ of the downstream classifier trained on noisy representations. If we steal the encoder using noisy representations and then train the downstream classifier using the stolen encoder, the accuracy is around 10\% worse than $A_L$. This is a good starting point but we need a better method for the defense to make a wider gap in terms of accuracy between legitimate and malicious users.

\begin{table}[t]
\caption{\textbf{Noise based perturbations} as a defence. The downstream accuracy on CIFAR10 for a legitimate user and an adversary are shown where the adversary used the \textbf{MSE} loss for stealing the representation model. All results are based on 9000 queries from the CIFAR10 test set. The noise added was Gaussian noise with a mean of \textbf{0} and standard deviation of $\sigma$ as in the table. \ahmad{To be removed / updated} }
\label{tab:compare-defence2}
\vskip 0.15in
\begin{center}
\begin{small}
\begin{sc}
\begin{tabular}{lccr}
\toprule
$\sigma$ \textbackslash Dataset & Legitimate & Adversary \\
\midrule
0 &  66.7 & 66.3 \\
1    & 63.9 & 64.8  \\
2  &  67.2 & 66.5 \\
3 & 65.0 & 65.5 \\
4   & 63.1 & 63.4 \\
\bottomrule
\end{tabular}
\end{sc}
\end{small}
\end{center}
\vskip -0.1in
\end{table}

\begin{table}[t]
\caption{\textbf{Noise based perturbations} as a defence where only very similar queries were perturbed by a large amount. The downstream accuracy on CIFAR10 for a legitimate user and an adversary are shown where the adversary used the \textbf{InfoNCE} loss (with two augmentations per image). All results are based on 9000 queries from the CIFAR10 test set. The noise added was Gaussian noise with a mean of 1000 and standard deviation of $\sigma$ as in the table. The first set of results are with the Cosine similarity as the distance metric and the second set of results is using the $\ell_2$ distance. }
\label{tab:compare-defencehigh}
\vskip 0.15in
\begin{center}
\begin{small}
\begin{sc}
\begin{tabular}{lccr}
\toprule
$\sigma$ \textbackslash Dataset & Legitimate & Adversary \\
\midrule
0  & 66.9  & 53.1  \\ 
20   & 65.4 & 41.2 \\
40      & 65.3 & 44.7 \\
60      & 66.0 & 41.8 \\
\midrule
0  & 66.9  & 52.2  \\ 
20   & 65.4 & 44.9 \\
40      & 65.3 & 45.7 \\
60      & 64.7 & 45.4 \\
\bottomrule
\end{tabular}
\end{sc}
\end{small}
\end{center}
\vskip -0.1in
\end{table}

\begin{table}[t]
\caption{\textbf{False Positive and False Negative} rates for different threshold values $\tau$ for the identification of similar queries. This identification is used by the defense which perturbs similar queries. The \textbf{$\ell_2$ distance} is used and the victim model passes the representation through its head after which the threshold $\tau$ is used. 
}
\label{tab:compare-threshold}
\vskip 0.15in
\begin{center}
\begin{small}
\begin{sc}
\begin{tabular}{lccr}
\toprule
$\tau$ & False Positive Rate (\%) & False Negative Rate (\%) \\
\midrule
12  & 4.7 $\pm$ 1.7 & 53.1 $\pm$ 2.8    \\ 
13   & 12.3 $\pm$ 3.1 & 35.1 $\pm$ 3.1  \\
14      &  27.7 $\pm$ 3.4  & 19.3 $\pm$ 2.2  \\
\bottomrule
\end{tabular}
\end{sc}
\end{small}
\end{center}
\vskip -0.1in
\end{table}

\begin{table}[t]
\caption{\textbf{False Positive and False Negative} rates for different threshold values $\tau$ for the identification of similar queries. This identification is used by the defense which perturbs similar queries. The \textbf{$\ell_2$ distance} is used and the victim model uses the representations directly. 
}
\label{tab:compare-threshold2}
\vskip 0.15in
\begin{center}
\begin{small}
\begin{sc}
\begin{tabular}{lccr}
\toprule
$\tau$ & False Positive Rate (\%) & False Negative Rate (\%) \\
\midrule
17   & 7.7 $\pm$ 2.3 & 49.0 $\pm$ 2.5   \\ 
18   & 16.2 $\pm$ 2.7 & 39.5 $\pm$ 2.5 \\
19      & 28.2 $\pm$ 2.6  & 26.4 $\pm$ 2.2 \\
\bottomrule
\end{tabular}
\end{sc}
\end{small}
\end{center}
\vskip -0.1in
\end{table}

\begin{table}[t]
\caption{\textbf{False Positive and False Negative} rates for different threshold values $\tau$ for the identification of similar queries. This identification is used by the defense which perturbs similar queries. \textbf{Cosine similarity} is used and the victim model passes the representations through the head before using the threshold. Note that with cosine similarity, a similarity value $> \tau$ is classified as similar. 
\ahmad{These tables are probably no longer important to refer to or talk about since a) the results are not fully up to date and b) we do not want to focus specifically on the types of defenses. }
}
\label{tab:compare-threshold3}
\vskip 0.15in
\begin{center}
\begin{small}
\begin{sc}
\begin{tabular}{lccr}
\toprule
$\tau$ & False Positive Rate (\%) & False Negative Rate (\%) \\
\midrule
0.45   & 31.4 $\pm$ 2.7  & 9.4 $\pm$ 1.8  \\
0.5   & 15.5 $\pm$ 2.6 & 21.3 $\pm$ 2.3    \\ 
0.55 & 7.0 $\pm$ 1.7  & 34.1 $\pm$ 2.1 \\
\bottomrule
\end{tabular}
\end{sc}
\end{small}
\end{center}
\vskip -0.1in
\end{table}


\begin{table}[t]
\caption{\textbf{Watermark Success Rate} on rotation prediction for models stolen with SVHN queries and different losses.}
\label{tab:watermarking}
\vskip 0.15in
\small
    \centering
        \begin{tabular}{lccc}
        \toprule
                      \textbf{\# Queries }  &  \textbf{ MSE } &  \textbf{ SoftNN } &  \textbf{ InfoNCE } \\
        \midrule
        10000 & 55.35 & 55.82 & 57.45  \\
        20000 & 55.92 & 55.32 & 57.2  \\
        30000 & 58.19 & 54.5 & 57.78  \\
        40000 & 55.78 & 56.68 & 55.75  \\
        50000 & 58.84 & 57.11 & 58.98 \\
        60000 & 56.84 & 57.58 & 58.02  \\
        70000 & 56.89 & 59.53 & 56.52 \\
        \bottomrule
        \end{tabular}
\end{table}

\begin{table}[t]
\caption{\textbf{Watermark Success Rate} on rotation prediction for models stolen with CIFAR10 queries and different augmentations applied to queries. \textbf{Standard} refers to the standard augmentations used in SimCLR, \textbf{+Rotation} refers to the standard augmentations along with rotations, and \textbf{NoAug} refers to completely unaugmented images.}
\label{tab:watermarking_cifar_aug}
\vskip 0.15in
\small
    \centering
        \begin{tabular}{lccccccccc}
        \toprule
                      \textbf{\# Queries }  & &  \textbf{ MSE } & & & \textbf{ SoftNN } & & & \textbf{ InfoNCE } \\
        \cmidrule(lr){1-1} \cmidrule(lr){2-4} \cmidrule(lr){5-7} \cmidrule(lr){8-10} 
                    & Standard & +Rotation & NoAug & Standard & +Rotation & NoAug & Standard & +Rotation & NoAug \\
        \midrule
        10000 & 54.69 & 73.76 & 54.61 & 55.48 & 74.49 & 56.45 & 56.53 & 56.85 & 57.0 \\
        20000 & 60.62 & 74.68 & 60.49 & 58.43 & 76.67 & 59.45 & 58.38 & 59.52 & 57.72 \\
        30000 & 61.45 & 75.84 & 61.45 & 59.03 & 77.86 & 62.03 & 58.01 & 59.92 & 58.84 \\
        40000 & 61.21 & 76.15 & 62.66 & 62.56 & 77.86 & 60.86 & 58.84 & 60.99 & 61.76 \\
        50000 & 59.81 & 76.5 & 60.95 & 62.15 & 78.53 & 63.94 & 59.42 & 63.21 & 62.34 \\
        \bottomrule
        \end{tabular}
\end{table}

\section{Hierarchy of Loss Functions}

\forestset{
sn edges/.style={for tree={parent anchor=south, child anchor=north,align=center,edge={->},base=bottom,where n children=0{}{}}}, 
background tree/.style={for tree={text opacity=0.2,draw opacity=0.2,edge={draw opacity=0.2}}}
}
\begin{figure}
\begin{center}
\begin{forest}
  [quadratic 
    [symmetrized]
    [contrastive
       [supervised contrastive]
    ]
  ]
\end{forest}
\end{center}
\caption{Hierarchy of the loss functions.}
\label{fig:loss-hierarchy}
\end{figure}
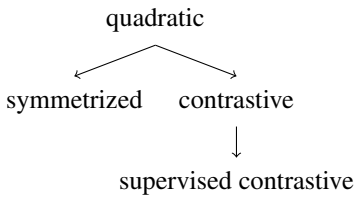

We present a hierarchy of loss functions in Figure~\ref{fig:loss-hierarchy}.

\section{Hierarchy of Defenses}
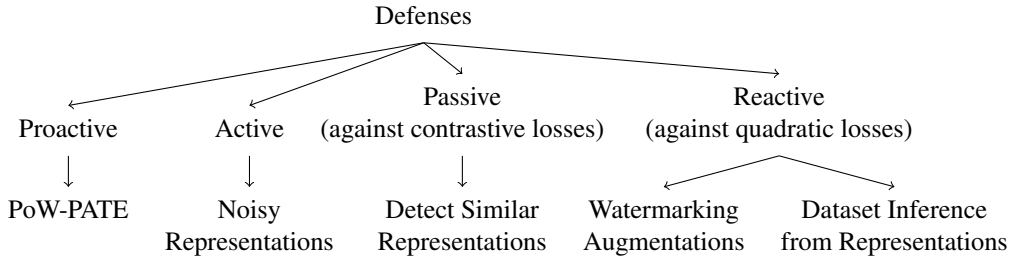
\begin{figure*}
\begin{center}
\begin{forest}
  [Defenses 
    [Proactive
        [PoW-PATE]
    ]
    [Active 
       [Noisy \\ Representations]
    ]
    [Passive \\ (against contrastive losses)
        [Detect Similar \\ Representations]
    ]
    [Reactive \\ (against quadratic losses)
        [Watermarking \\ Augmentations]
        [Dataset Inference \\ from Representations]
    ]
  ]
\end{forest}
\end{center}
\caption{Hierarchy of the defense methods.}
\label{fig:defense-hierarchy}
\end{figure*}

A hierarchy of defense methods is presented in Figure~\ref{fig:defense-hierarchy}.

\section{Stealing Algorithms} 

\begin{algorithm}
  \caption{Training a Downstream Classifier for Linear Evaluation Protocol}\label{alg:downstream}
  
  \algorithmicrequire{Victim or Stolen Model $f$, downstream task with labeled data}
  
  \algorithmicensure{Downstream linear classifier $g(x; \theta)$} 

   \begin{algorithmic}[1]
   \STATE Add a linear layer to the model $f$ and freeze all previous layers
   \FOR{labeled query batches $\{x_k, y_k\}_{k=1}^{N}$}
    \STATE Compute loss $\mathcal{L} \left\{ \{f(x_k), y_k\} \right\}$ where $\mathcal{L} \{f(x),y\} = - \sum_{i} y_k \log f(x)$
    \STATE Update linear model parameters $\theta \coloneqq \theta - \eta \nabla_{\theta} \mathcal{L}$
    \ENDFOR
    
  \end{algorithmic}
\end{algorithm}

\section{Defenses}

\subsection{Prediction Poisoning}
\label{sec:prediction-poisoning-analysis}

We can consider the gradient of the loss with respect to the model parameters $\theta \in \mathbb{R}^{D}$ for a query $x$ made by the attacker to formalize this objective as in~\cite{orekondy2019prediction}. We let $y_v \in \mathbb{R}^n$ be the representations returned by the victim model and $\tilde{y_v}$ be the perturbed representations which will act as targets for the stolen model $F$ to be trained by the attacker. Then the gradient of the loss function which will be used by the attacker to update its model parameters based on the unperturbed representations is $a = -\nabla_{\theta} \mathcal{L} \{ F(x; \theta), y_v \}$. Similarly, let $b = -\nabla_{\theta} \mathcal{L} \{ F(x; \theta), \tilde{y_v} \}$ be the gradient when the perturbed representations are returned. To harm the attacker's gradient, the similarity between $a$ and $b$ should be minimized. This similarity can be quantified with various metrics such as the $\ell_2$ distance or cosine similarity. We let $G$ be the downstream linear classifier trained by a legitimate user and $\phi \in \mathbb{R}^E$ be the parameters of this model. Further, let $J$ be the loss function used to train this classifier and $t \in \mathbb{R}^k$ be the target labels from the dataset used for the downstream task. Then in order to avoid harming the legitimate user, the victim must maximize the similarity between 
\begin{equation*}
    c = -\nabla_{\phi} J \{ G(y_v; \phi), t \}
\end{equation*} 
and 
\begin{equation*}
d = -\nabla_{\phi} J \{ G(\tilde{y_v}; \phi), t \}
\end{equation*}
Although there are various loss functions $\mathcal{L}$ which can be used by an attacker when training a stolen model, we assume for simplicity that the attacker uses the Mean Squared Error (MSE) loss. For the downstream task which involves a standard training task, we assume standard training so that $J$ can be chosen to be the Cross Entropy Loss.  With these assumptions, $a$ can be simplified as:
\begin{align*}
    & -\nabla_{\theta} \mathcal{L} \{ F(x; \theta), y_v \} \\
    & =  -\nabla_{\theta} \frac{1}{n} \sum_{i=1}^{n} (F(x; \theta)_i - y_{v_i})^2 \\
    & = - \frac{1}{n} \sum_{i=1}^{n} \nabla_{\theta} (F(x; \theta)_i - y_{v_i})^2  \\
    & = - \frac{1}{n} \sum_{i=1}^{n} 2 (F(x; \theta)_i - y_{v_i}) \nabla_{\theta} (F(x; \theta)_i - y_{v_i})  \\
    & = -\frac{2}{n} \sum_{i=1}^{n} (F(x; \theta)_i - y_{v_i}) \cdot \nabla_{\theta} (F(x; \theta)_i)
\end{align*}
since $\nabla_{\theta} y_{v} = 0$. This can then be simplified as the matrix product 
\begin{equation*}
    a = -\frac{2}{n} \nabla_{\theta} F^{T} (F(x; \theta)- y_v)
\end{equation*}
 where $\nabla_{\theta} F \in \mathbb{R}^{n \times D}$ is the Jacobian of the stolen model $F$.  In a similar way, $b$ can be simplified to 
 \begin{equation*}
     b = -\frac{2}{n} \nabla_{\theta} F^{T} (F(x; \theta)-\tilde{y_v})
 \end{equation*}
 To optimize $\tilde{y_v}$ so that the similarity between $a$ and $b$ is minimized, we thus need to recreate the attacker's model $F$ in some form to approximate the Jacobian $ \nabla_{\theta} F$ and the model's output $F(x)$. This can be done by using a surrogate model $F_{surr}$ on the victim's end. Then if we let $\mathrm{sim}$ be the similarity function used, the optimization objective can be written as 
 
 \begin{equation*}
     \min_{\tilde{y_v}} \mathrm{sim} (a,b) = \min_{\tilde{y_v}} \mathrm{sim} (-\frac{2}{n} \nabla_{\theta} F_{surr}^{T} (F_{surr}(x; \theta)-y_v), -\frac{2}{n} \nabla_{\theta} F_{surr}^{T} (F_{surr}(x; \theta)-\tilde{y_v}))
 \end{equation*}

We also simplify $c$ as 

\begin{align*}
    & -\nabla_{\phi} J \{ G(y_v; \phi), t \}  \\
    &= -\nabla_{\phi} -\sum_{i=1}^{k} t_i \log G(y_v; \phi)_i \\
    &= \sum_{i=1}^{k} \nabla_{\phi} t_i \log G(y_v; \phi)_i  \\
    &= \sum_{i=1}^{k} t_i \nabla_{\phi} \log G(y_v; \phi)_i \\
    &= \nabla_{\phi} \log G(y_v)^T t \\
\end{align*}

and $d$ as   $\nabla_{\phi} \log G(\tilde{y_v})^T t$. In this case, we note that to satisfy this optimization requirement, the victim must recreate the downstream classifier $G$ to estimate the Jacobians $\nabla_{\phi} \log G(y_v; \phi) \in \mathbb{R}^{k \times E}$ and $\nabla_{\phi} \log G(\tilde{y_v}; \phi)$. Again this can be done using a surrogate model $G_{surr}$ on the victim's end based on which we can write the optimization objective for the downstream task as 
\begin{equation*}
    \max_{\tilde{y_v}} \mathrm{sim} (\nabla_{\phi} \log G_{surr}(y_v)^T t, \nabla_{\phi} \log G_{surr}(\tilde{y_v})^T t)
\end{equation*}
 We note that different similarity functions may be used for the two objectives. Therefore applying prediction poisoning requires a double optimization problem to be solved, each of which requires certain assumptions such as the type of loss function used.



\subsection{Watermarking Defense}
We include additional results for our Watermarking defense which embeds predictive information about a particular augmentation (rotations, in our case) in the features learned by the victim. The ability of the watermark to transfer is measured by the Watermark Success Rate, which is the accuracy of a classifier trained to predict the watermarking augmentation. In our experiments, this task is prediction between images rotated by angles in $[0^{\circ}, 180^{\circ}]$ or $[180^{\circ}, 360^{\circ}]$. A random, genuinely but separately trained, model achieves random performance on this task, i.e. around 50\% watermark success rate. Table~\ref{tab:watermarking} shows that the watermark transfers sufficiently for models stolen with SVHN queries and different losses. Since the watermarking strategy is closely related to the augmentations used during training or stealing, we test the watermark success rate for CIFAR10 queries that are augmented with the standard SimCLR augmentations, standard augmentations with rotations included, or no augmentations at all. We report these results in Table~\ref{tab:watermarking_cifar_aug} and find that in all cases, the watermark transfers to the stolen model well. If the adversary happens to use rotations as an augmentation, the watermark transfer is significantly stronger. However, this is not necessary for good watermark transfer, as demonstrated by the NoAug success rates.

\subsection{Dataset Inference Defense} 

We also tested an approach based on Dataset Inference~\citep{maini2021dataset} as a possible defense. In the case of representation models, we directly measure the $\ell_2$ distance between the representations of a victim model and a model stolen with various methods. We also compute the distance between the representations of a victim model and a random model, which in this case was trained on the same dataset and using the same architecture but a different random seed that resulted in different parameters. The distance is evaluated on the representations of each model on the training data used to train the victim model. 
We obtain the results averaged across the datasets in Table~\ref{tab:datasetinfBig}. The general trend is that the more queries the adversaries issue against the victim model that are used to train the stolen copy, the smaller the distance between the representations from the victim and stolen models. In terms of loss functions, the stealing with MSE always results in smaller distances than for another Random model. On the other hand, for attacks that use either InfoNCE or SoftNN losses, it is rather hard to differentiate between stolen models and a random model based on the distances of their representations from the victim as some have a higher distance while others have a lower distance. 
We also observe a similar trend with watermarks where the MSE and SoftNN loss methods give different performances compared to a random model.

The more an adversary interacts with the victim model to steal it, the easier it is to claim ownership by distinguishing the stolen model’s behavior on the victim model’s training set.

\begin{table*}[t]
\caption{$\ell_2$ distances between the representation of a stolen model and a victim model. The Random column denotes model of the same architecture and training procedure like the victim model but with a different initial random seed. The Victim column designates the type of dataset used by the victim. The AdvData represents the data used by an adversary to steal the victim model.\adam{We should use victim's training set to measure these distances.}}
\label{tab:datasetinfBig}
\vskip 0.15in
\begin{center}
\begin{small}
\begin{sc}
\begin{tabular}{ccc|cccc}
\toprule
\# Queries & Victim & AdvData & Random & MSE & InfoNCE & SoftNN \\
\midrule
9K & CIFAR10 & CIFAR10 & 34.1 & 26.6 & 44 & 14.4 \\
50K & CIFAR10 & CIFAR10 & 34.1 & 14.6 & 40 & 12.8 \\
50K & ImageNet & CIFAR10 & 65.8 & 25.2 & 36.2 & 119.6 \\
\bottomrule
\end{tabular}
\end{sc}
\end{small}
\end{center}
\vskip -0.1in
\end{table*}


\begin{table*}[t]
\caption{\textbf{Dataset Inference.} 
Compare the differences between raw data points and their augmentations for different private and public datasets. We use statistical t-test where the null hypothesis is that the differences in distances are smaller for the public dataset than for the private dataset, where $\Delta \mu$ is the effect size. \textit{Size} is the image size. U is un-/self-supervised learning, while S denotes the supervised learning.}
\label{tab:dataset-inference-main}
\vskip 0.15in
\begin{center}
\begin{small}
\begin{sc}
\begin{tabular}{cccccccccc}
\toprule
Size & \# Queries & \# Aug. & Model & Learning & Private & Public & $\Delta \mu$ & p-value & t-value \\
\midrule
32 & 100 & 100 & SVHN & U & SVHN & SVHN & -0.17 & 0.99 & -2.90 \\
32 & 10000 & 100 & SVHN & U & SVHN & SVHN & -0.005 & 0.79 & -0.82 \\
224 & 64 & 10 & ImageNet & U & ImageNet & ImageNet & -0.005 & 0.99 & -2.53 \\
224 & 32 & 10 & ImageNet & S & ImageNet & ImageNet & 0.71 & 0.028 & 1.92 \\
224 & 64 & 10 & ImageNet & S & ImageNet & ImageNet & 1.06 & $10^{-5}$ & 4.03 \\
224 & 64 & 10 & ImageNet & S & ImageNet & ImageNet & 0.89 & $10^{-4}$ & 3.61 \\
224 & 128 & 10 & ImageNet & S & ImageNet & ImageNet & 0.36 & 0.02 & 1.97 \\
224 & 256 & 10 & ImageNet & S & ImageNet & ImageNet & -0.44 & 0.99 & -3.31 \\
224 & 256 & 10 & ImageNet & S & ImageNet & ImageNet & -0.12 & 0.81 & -0.87 \\
224 & 1024 & 10 & ImageNet & U & ImageNet & ImageNet & -0.02 & 1.0 & -33.99 \\
\hline
32 & 50 & 10 & SVHN & S & SVHN & SVHN & 0.15 & 0.0685 & 1.49 \\
32 & 50 & 10 & SVHN & S & SVHN & SVHN & 0.39 & 0.033 & 1.84 \\
32 & 50 & 10 & CIFAR10 & S & CIFAR10 & CIFAR10 & -0.46 & 0.99 & -4.3 \\
\bottomrule
\end{tabular}
\end{sc}
\end{small}
\end{center}
\vskip -0.1in
\end{table*}

We present results for the dataset inference method in Table~\ref{tab:dataset-inference-main}.

\newpage

\section {Representation Model Architecture} \label{app:modelarch}

This section shows a sample representation model architecture (in this case a ResNet34 model). (generated using \textit{torchsummary}).

\begin{Verbatim}
----------------------------------------------------------------
        Layer (type)               Output Shape         Param #
================================================================
            Conv2d-1           [-1, 64, 32, 32]           1,728
       BatchNorm2d-2           [-1, 64, 32, 32]             128
            Conv2d-3           [-1, 64, 32, 32]          36,864
       BatchNorm2d-4           [-1, 64, 32, 32]             128
            Conv2d-5           [-1, 64, 32, 32]          36,864
       BatchNorm2d-6           [-1, 64, 32, 32]             128
        BasicBlock-7           [-1, 64, 32, 32]               0
            Conv2d-8           [-1, 64, 32, 32]          36,864
       BatchNorm2d-9           [-1, 64, 32, 32]             128
           Conv2d-10           [-1, 64, 32, 32]          36,864
      BatchNorm2d-11           [-1, 64, 32, 32]             128
       BasicBlock-12           [-1, 64, 32, 32]               0
           Conv2d-13           [-1, 64, 32, 32]          36,864
      BatchNorm2d-14           [-1, 64, 32, 32]             128
           Conv2d-15           [-1, 64, 32, 32]          36,864
      BatchNorm2d-16           [-1, 64, 32, 32]             128
       BasicBlock-17           [-1, 64, 32, 32]               0
           Conv2d-18          [-1, 128, 16, 16]          73,728
      BatchNorm2d-19          [-1, 128, 16, 16]             256
           Conv2d-20          [-1, 128, 16, 16]         147,456
      BatchNorm2d-21          [-1, 128, 16, 16]             256
           Conv2d-22          [-1, 128, 16, 16]           8,192
      BatchNorm2d-23          [-1, 128, 16, 16]             256
       BasicBlock-24          [-1, 128, 16, 16]               0
           Conv2d-25          [-1, 128, 16, 16]         147,456
      BatchNorm2d-26          [-1, 128, 16, 16]             256
           Conv2d-27          [-1, 128, 16, 16]         147,456
      BatchNorm2d-28          [-1, 128, 16, 16]             256
       BasicBlock-29          [-1, 128, 16, 16]               0
           Conv2d-30          [-1, 128, 16, 16]         147,456
      BatchNorm2d-31          [-1, 128, 16, 16]             256
           Conv2d-32          [-1, 128, 16, 16]         147,456
      BatchNorm2d-33          [-1, 128, 16, 16]             256
       BasicBlock-34          [-1, 128, 16, 16]               0
           Conv2d-35          [-1, 128, 16, 16]         147,456
      BatchNorm2d-36          [-1, 128, 16, 16]             256
           Conv2d-37          [-1, 128, 16, 16]         147,456
      BatchNorm2d-38          [-1, 128, 16, 16]             256
       BasicBlock-39          [-1, 128, 16, 16]               0
           Conv2d-40            [-1, 256, 8, 8]         294,912
      BatchNorm2d-41            [-1, 256, 8, 8]             512
           Conv2d-42            [-1, 256, 8, 8]         589,824
      BatchNorm2d-43            [-1, 256, 8, 8]             512
           Conv2d-44            [-1, 256, 8, 8]          32,768
      BatchNorm2d-45            [-1, 256, 8, 8]             512
       BasicBlock-46            [-1, 256, 8, 8]               0
           Conv2d-47            [-1, 256, 8, 8]         589,824
      BatchNorm2d-48            [-1, 256, 8, 8]             512
           Conv2d-49            [-1, 256, 8, 8]         589,824
      BatchNorm2d-50            [-1, 256, 8, 8]             512
       BasicBlock-51            [-1, 256, 8, 8]               0
           Conv2d-52            [-1, 256, 8, 8]         589,824
      BatchNorm2d-53            [-1, 256, 8, 8]             512
           Conv2d-54            [-1, 256, 8, 8]         589,824
      BatchNorm2d-55            [-1, 256, 8, 8]             512
       BasicBlock-56            [-1, 256, 8, 8]               0
           Conv2d-57            [-1, 256, 8, 8]         589,824
      BatchNorm2d-58            [-1, 256, 8, 8]             512
           Conv2d-59            [-1, 256, 8, 8]         589,824
      BatchNorm2d-60            [-1, 256, 8, 8]             512
       BasicBlock-61            [-1, 256, 8, 8]               0
           Conv2d-62            [-1, 256, 8, 8]         589,824
      BatchNorm2d-63            [-1, 256, 8, 8]             512
           Conv2d-64            [-1, 256, 8, 8]         589,824
      BatchNorm2d-65            [-1, 256, 8, 8]             512
       BasicBlock-66            [-1, 256, 8, 8]               0
           Conv2d-67            [-1, 256, 8, 8]         589,824
      BatchNorm2d-68            [-1, 256, 8, 8]             512
           Conv2d-69            [-1, 256, 8, 8]         589,824
      BatchNorm2d-70            [-1, 256, 8, 8]             512
       BasicBlock-71            [-1, 256, 8, 8]               0
           Conv2d-72            [-1, 512, 4, 4]       1,179,648
      BatchNorm2d-73            [-1, 512, 4, 4]           1,024
           Conv2d-74            [-1, 512, 4, 4]       2,359,296
      BatchNorm2d-75            [-1, 512, 4, 4]           1,024
           Conv2d-76            [-1, 512, 4, 4]         131,072
      BatchNorm2d-77            [-1, 512, 4, 4]           1,024
       BasicBlock-78            [-1, 512, 4, 4]               0
           Conv2d-79            [-1, 512, 4, 4]       2,359,296
      BatchNorm2d-80            [-1, 512, 4, 4]           1,024
           Conv2d-81            [-1, 512, 4, 4]       2,359,296
      BatchNorm2d-82            [-1, 512, 4, 4]           1,024
       BasicBlock-83            [-1, 512, 4, 4]               0
           Conv2d-84            [-1, 512, 4, 4]       2,359,296
      BatchNorm2d-85            [-1, 512, 4, 4]           1,024
           Conv2d-86            [-1, 512, 4, 4]       2,359,296
      BatchNorm2d-87            [-1, 512, 4, 4]           1,024
       BasicBlock-88            [-1, 512, 4, 4]               0
         Identity-89                  [-1, 512]               0
           ResNet-90                  [-1, 512]               0
================================================================
Total params: 21,276,992
Trainable params: 21,276,992
Non-trainable params: 0
----------------------------------------------------------------
Input size (MB): 0.01
Forward/backward pass size (MB): 19.07
Params size (MB): 81.17
Estimated Total Size (MB): 100.25
----------------------------------------------------------------
\end{Verbatim}

\end{document}